\documentclass{article}
\usepackage[preprint]{nips_2018}
\usepackage[utf8]{inputenc} 
\usepackage[T1]{fontenc}    
\usepackage{hyperref}       
\usepackage{url}            
\usepackage{booktabs}       
\usepackage{amsfonts}       
\usepackage{nicefrac}       
\usepackage{microtype}      
\usepackage{multirow}
\usepackage{graphicx}
\usepackage{amsmath,amssymb}
\usepackage{algorithm}
\usepackage[noend]{algpseudocode}

\newcommand{\argmax}{\arg\!\max}

\begin{document}

\title{Expressive power of outer product manifolds on feed-forward neural networks}
 
\author{B\'alint Dar\'oczy \\
Institute for Computer Science and Control \\
Hungarian Academy of Sciences (MTA SZTAKI)\\
H-1111, Kende str. 13-17, Budapest, Hungary \\
\texttt{daroczyb@ilab.sztaki.hu} \\
\And 
Rita Aleksziev \\
Institute for Computer Science and Control \\
Hungarian Academy of Sciences (MTA SZTAKI)\\
H-1111, Kende str. 13-17, Budapest, Hungary \\
\texttt{alekszievr@ilab.sztaki.hu} \\
\AND
Andr\'as Bencz\'ur\\
Institute for Computer Science and Control \\
Hungarian Academy of Sciences (MTA SZTAKI)\\
H-1111, Kende str. 13-17, Budapest, Hungary \\
\texttt{benczur@ilab.sztaki.hu} \\
}

\maketitle

\begin{abstract}
Hierarchical neural networks are exponentially more efficient than their corresponding ``shallow'' counterpart with the same expressive power, but involve huge number of parameters and require tedious amounts of training.
Our main idea is to mathematically understand and describe the hierarchical structure of feedforward neural networks by reparametrization invariant Riemannian metrics. By computing or approximating the tangent subspace, we better utilize the original network via sparse representations that enables switching to shallow networks after a very early training stage. Our experiments show that the proposed approximation of the metric improves and sometimes even surpasses the achievable performance of the original network significantly even after a few epochs of training the original feedforward network.
\end{abstract}


\section{Introduction}

Hierarchical neural networks are among the most important machine learning models and are deemed to be the state-of-the-art models in problems like image classification \cite{szegedy2016rethinking,he2016deep}, various natural language processing problems \cite{greff2017lstm}, object detection \cite{ren2015faster}, image captioning \cite{vinyals2017show} or reinforcement learning \cite{silver2017mastering}. There are several challenges that are related to the generalizational power, the expressive power \cite{bartlett2003vapnik,zhang2016understanding,lin2016does,rolnick2017power,bengio2011expressive,kirkpatrick2017overcoming}, the efficiency,  \cite{szegedy2016rethinking,he2016deep,greff2017lstm,kingma2013auto,goodfellow2014generative,ren2015faster} and the optimization methods \cite{ioffe2015batch,srivastava2014dropout,amari1996neural,kingma2014adam,ollivier2015riemannian} of these models.  



Inference for feed-forward neural networks with common, almost everywhere continuously differentiable cost and activation functions are usually trainable via the partial derivatives of the loss function respect to the parameters of the network \cite{rumelhart1985learning}, using some sort of regularization method \cite{kingma2014adam,zeiler2012adadelta,srivastava2014dropout,ioffe2015batch}. As a common choice, training algorithms use back-propagation on mini batches to estimate the first and second order gradient \cite{nesterov1983method,amari1996neural,kingma2014adam}. The non-convex nature of the loss functions of feed-forward neural networks makes simple gradient based learning methods result in convergence to a local minimum instead of a global one. There are some refined ideas to overcome this serious issue. For example DropOut \cite{srivastava2014dropout} may make the model ``jump'' out of local minima with non-zero probability, although at the same time it can make it ``jump'' away from the global minimum too.  The second order derivatives form the Hessian matrix \cite{shima1995hessian,amari2014curvature} and are used as the normalization part of the ``Newtonian'' gradient. 

In the paper, we investigate the connection between the structure of a neural network and Riemannian manifolds to gain a better understanding and to utilize more of their potential. In a way, many of the existing machine learning problems can be investigated as statistical learning problems. Although information geometry \cite{amari1996neural} plays an important role in statistical learning, the geometrical properties of target functions both widely used and recently discovered, along with those of the models themselves, are not well studied. 

Over the parameter space and the error function we can often determine a smooth manifold \cite{ollivier2015riemannian}. In this paper we investigate the tangent bundle of this manifold in order to understand the behavior of certain networks better and to take advantage of specific Riemannian metrics having unique invariance properties \cite{Cencov1982,campbell1986extended}. This approach may tell us more about what a neural network can and cannot express \cite{lin2016does}. We aim to understand how the chosen learning algorithm, target function and the complexity of the chosen model determine the geometrical properties of the underlying manifolds. This knowledge might allow us to determine the achievable approximation and estimation error of a model before finishing its training.

We consider various smooth manifolds on the parameter space derived from the error function and examine some Hessian based Riemannian metrics. We use the partial derivatives in the tangent space as representation of data points. The inner products in the tangent space are quadratic, therefore if we separate the samples with a second order polynomial, then the actual metric will be irrelevant. 

Our contributions are the following.

\begin{itemize}
\item We prove that a class of Riemannian metrics called the outer product metrics on feed-forward discriminative and generative neural networks are invariant to reparametrization and the inner product is sub quadratic. Hence the inner product can be used to optimize a weakly trained NN.
\item We describe an approximation algorithm for the inner product space, in case of weekly trained NNs of various sparsities.
\item We give heuristics for classification tasks where the coefficients of the monomials in the inner product are not necessarily determined.
\end{itemize}

Our experiments were done on the CIFAR-10 \cite{krizhevsky2009learning} and MNIST \citep{lecun-mnisthandwrittendigit-2010} data sets. We showed that if we determine the extended Hessian manifold in an early stage of learning of the underlying feed-forward neural network (after only a few epochs), we can outperform the fully trained network by passing the linearized inner product space to a shallow network. 

\section{Related work}

Feed-forward neural networks are described in \cite{bishop1992exact,zhang2016understanding,rolnick2017power,lin2016does}. Bengio et al. consider a feed-forward neural network deep if the number of hidden layers is more than one \cite{bengio2011expressive}. The most common layer types are the fully connected and the convolutional layer while the most popular activation functions (which are the non-linear functions applied to the linear combination of the preceding layer's activations), are sigmoid, hyperbolic tangent (tanh) or some variant of the rectified linear unit (ReLU). In case of discriminative models, the activation function of the output layer is usually softmax. 

To determine the parameters of a network, some form of gradient descent is used. Back-propagation was introduced in \cite{rumelhart1985learning} and even simple regularization \cite{krogh1992simple} plays an important role during training. The problem of ``vanishing gradients'' became a serious issue in recurrent neural networks \cite{greff2017lstm} but with the deeper feed-forward neural networks the issue became relevant again. As a result, sigmoid and tanh is commonly replaced by ReLU because of the absent upper bound of its derivative. Ioffe and Szegedy in \cite{ioffe2015batch} suggested normalization of the input of the non-linear functions to prevent ReLU from outputting zero. Techniques mentioned so far do not violate the smoothness property we will count on. 

In a promising theoretical result \cite{lin2016does}, the authors inspect the expressive power  of shallow neural networks, stating that efficient ``flattening'' of deep architectures is exponentially expensive even for simple cases. In \cite{rolnick2017power} the authors show the quantification of expressive power of deeper ANNs and define lower bounds for ``deepness'' to approximate a given function. The result is in accordance with \cite{bengio2011expressive} without considering bounds for convergence as in \cite{zhang2016understanding}. These results and the general results about \textit{VC-dimension} of neural networks \cite{sontag1998vc,bartlett2003vapnik} raise questions about the approximation achievable by the structure and the transformations in hierarchical models. 

The geometrical property of the underlying manifold was used for optimizing generative models \cite{rifai2011manifold} and as a general framework for optimization in \cite{ollivier2015riemannian,zhang2016riemannian}, neither of them utilize the tangent space as representation. The closest to our method is \cite{jaakkola1999exploiting} where the authors used the diagonal of the Fisher information matrix of some generative probability density functions in a kernel function for classification.  Closed formula for Gaussian Mixtures was proposed in \cite{perronnin2007fisher}. 

\section{Outer product manifolds of feed-forward networks}

Feed-forward networks with parameters $\theta$ solve the optimization problem
\begin{equation}
\label{eq:expectation}
\min_\theta f(\theta) = \mathbb{E}_{\mathcal{X}}[l(x;\theta)],
\end{equation}
where $l(x;\theta)$ is usually a non-convex function of $\theta$. In case of discriminative models the loss function depends on the target variable as well: $l(x;c,\theta)$. We define a Riemannian manifold $(\mathcal{M},g)$ based on (\ref{eq:expectation}) by assigning a tangent subspace $T_{\theta}M$ to each configuration point $\theta$ with an inner product via a Riemannian metric $g_{\theta}: T_{\theta}M \times T_{\theta}M \rightarrow \mathbb{R}$ where $\theta \in \Theta \subset \mathcal{M}$. If we minimize over a finite set of known examples, then the problem is closely related to the empirical risk minimization and loglikelihood maximization. 

The parameter space of continuously differentiable feed-forward neural networks (CDFNN) has a Riemannian metric structure \cite{ollivier2015riemannian}. Formally, let $X=\{x_1,..,x_T\}$ be a finite set of known observations with or without a set of target variables $Y=\{y_1,..,y_T\}$ and a directed graph $N=\{V,E\}$ where $V$ is the set of nodes with their activation functions and $E$ is the set of weighted, directed edges between the nodes. Let the loss function $l$ be additive over $X$. Now, in case of generative models, the optimization problem has the form $\min_{f \in \mathcal{F}_N} l(X;f) = \min_{f \in \mathcal{F}_N} \frac{1}{T}\sum_i^T l(x_i;f)$ where $\mathcal{F}_{N}$ is the class of neural networks with structure $N$. 

To determine a suitable model $N$, we need an optimization algorithm to search through the possible candidate networks. Inference for finite feed-forward neural networks with common, almost everywhere continuously differentiable cost and activation functions is usually trainable via back-propagation \cite{amari1996neural,rumelhart1985learning}, a method based on partial derivatives. Optimization can be interpreted as a ``random walk'' on the manifold with finite steps defined by some transition function between the points and their tangent subspaces. The fundamental theorem of Riemannian geometry states that there is a unique transition between two points, the Levi-Civita connection $\nabla$ which preserves the metric from point to point on the manifold. 

The general constraint about Riemannian metrics is that the metric tensor should be symmetric, positive definite and the inner product in the tangent space assigned to any point on a finite dimensional manifold has to be able to be formalized as $<x,x>_{\theta} = dx^T G_{\theta} dx = \sum_{i,j} g_{i,j}^{\theta} dx^i dx^j$. The metric $g_{\theta}$ varies smoothly by $\theta$ on the manifold and is arbitrary given the conditions. In this paper we will be focusing on metrics which are not very sensitive to invertible changes and preserve the inner product and we suggest an optimization method on the tangent space based on a kernel function.

\subsection{Outer product manifolds}

Let our loss function $l(x;\theta)$ be a smooth, positive, parametric real function where $x \in \mathbb{R}^d$ and $\theta \in \mathbb{R}^n$. We define a class of $n \times n$ positive semi-definite matrices as 

\begin{equation}
\label{eq:LC_dot}
h_{\theta}(x) = \nabla_{\theta} l(x;\theta) \otimes \nabla_{\theta} l(x;\theta)
\end{equation}

where $\nabla$ is the Levi-Civita connection as $ \nabla_{\theta} l(x;\theta) = \{\frac{\partial l(x;\theta)}{\partial \theta_1},..,\frac{\partial l(x;\theta)}{\partial \theta_n}\}$. 
%
%
%
Using eq.~(\ref{eq:LC_dot}) we can determine a class of Riemannian metrics 
\begin{equation}
G = g_{X}(h_{\theta}(x))
\label{eq:riemann-metric}
\end{equation}
where $g_X$ is a quasi arithmetic mean over $X$. For example, if $g_X$ is the arithmetic mean, then the metric is $ G_{\theta}=\mathbf{AM}_{X} [\nabla _\theta l(x_i|\theta)\nabla _\theta l(x_i|\theta)^T]$ and we can approximate it with a finite sum as $G_{\theta}^{kl} \approx \sum_{i} \omega_i
\left(\frac{\partial}{\partial \theta_k} l(x_i|\theta)\right)
\left(\frac{\partial}{\partial \theta_l} l(x_i|\theta)\right)$ with some importance $\omega_i$ assigned to each sample. Through $G$, the tangent bundle of the Riemannian manifold induces a normalized inner product (kernel) at any configuration of the parameters formalized for two samples $x_i$ and $x_j$ as 

\begin{equation}
\label{eq:K_base}
<x_i,x_j>_{\theta} = \nabla_{\theta} l(x_i;\theta)^T G_{\theta}^{-1} \nabla_{\theta} l(x_j;\theta)
\end{equation}

where the inverse of $G_{\theta}$ is positive semi-definite since $G_{\theta}$ is positive semi-definite.  

\textbf{Proposition 1:}
These kernel functions satisfy the Mercer conditions and are reparametrization invariant: any invertible, continuously differentiable change $\rho$ of the parameters keeps the kernel, meaning that for $\theta = \rho(\mu)$, $<,>_\mu$ is identical to $<,>_\theta$ . 

\textbf{Proof:}
The partial derivates at $\mu$ are
\begin{equation}
\nonumber
\nabla(\mu) = \nabla(\rho(\mu)) \bigg(\frac {\partial\rho}{\partial \mu}\bigg) 
\end{equation}
and therefore 
\begin{gather*}
<,>_{\mu} = \nabla_{\mu}^T G_{\mu}^{-1} \nabla_{\mu} \\ 
= \nabla_{\rho(\mu)} \bigg(\frac {\partial\rho}{\partial \mu}\bigg)^T \Bigg(G_{\rho(\mu)} \bigg(\frac {\partial\rho}{\partial \mu}\bigg)^2 \Bigg)^{-1} \nabla_{\rho(\mu)}\bigg(\frac {\partial\rho}{\partial \mu}\bigg) \\
= \nabla_{\rho(\mu)}^T G_{\rho(\mu)}^{-1}  \nabla_{\rho(\mu)} = <,>_{\rho}.
\square
\end{gather*}

The quadratic nature of the Riemannian metrics is a serious concern due to high dimensionality of the tangent space. By Nash's embedding theorem any finite dimensional Riemannian manifold can be embedded into a usually higher, but finite dimensional Euclidean space via an isometry. There are several ways to determine a linear inner product: decomposition or diagonal approximation of the metric, or quadratic flattening. Due to high dimensionality, both decomposition and flattening can be highly inefficient, although flattening can be highly sparse in case of sparsified gradients. By flattening we consider the inner product as a normalized linear space, $<,>= \sum_{i,j} g_{i,j} \frac{\partial}{\partial \theta_i} \frac{\partial}{\partial \theta_j} = \sum_i ( \hat{g}_{i} \frac{\partial}{\partial \theta_i})^T \hat{g}_{i} \frac{\partial}{\partial \theta_i}$ where $\hat{g}_i = \sum_j g_{i,j} \frac{\partial}{\partial \theta_j}$ can be determined via decomposition and assigning a flattened vector $\{\hat{g}_{0}\frac{\partial}{\partial \theta_{0}},\hat{g}_{0}\frac{\partial}{\partial \theta_{1}},\hat{g}_{0}\frac{\partial}{\partial \theta_{2}},..,\hat{g}_{n}\frac{\partial}{\partial \theta_{n}}\}$. Now, let the sparsity $s(x)$ be the proportion of the nonzero elements in gradient vector $x$ then the sparsity of the resulted quadratic will be at most $s(x)^2$. 

Next, we consider Markov Random Fields with log-likelihood and some discriminative feed-forward neural networks. 

\subsection{Outer product space and generative Markov Random Fields}

Besides the invariance property of the outer product metric, if our loss function is $l(x;\theta)=\log p(x;\theta)$ where $p$ is a parametric probability density function of a generative model, the outer product metric will be the Fisher information matrix, $F_{\theta}=\mathbf{E_{p(x;\theta)}[\nabla_{\theta} l(x_i;\theta) \otimes \nabla_{\theta} l(x_j;\theta)]}$ with a unique invariance property under \textit{Markov morphisms} \cite{Cencov1982,campbell1986extended}. 

Let our generative model be a Markov Random Field (MRF) then by the Hammersley-Clifford theorem \cite{hammersley1971markov}, the distribution is  a Gibbs distribution, which can be factorized over the maximal cliques and expressed by a potential function $U$ over the maximal cliques $C=\{c_k\}$ as $p(x;\theta) = {\mathrm{e}^{-U(x;\theta)}}/{Z(\theta)}$ where $U(x;\theta) = \sum_{c_k \in C} u_{k}(x;\theta)$ is the energy function and $Z(\theta) = \sum_{i} \mathrm{e}^{-U(x;\theta)}$ is the sum of the exponent of the energy function over our generative model, a normalization term called the partition function. If the model parameters are previously determined, then $Z(\theta)$ is a constant and we can prove the following:

\textbf{Proposition 2:}
For MRF the Fisher information is
\begin{equation}
\begin{split}
G_{\theta}^{k,l} &={\bf E_{\theta}}[\nabla_{\theta_k} \log p(x;\theta) \nabla_{\theta_l} \log p(x;\theta)] \\
&= {\bf E_{\theta}}[({\bf E_{\theta}}[\frac{\partial{u_k(x;\theta)}}{\partial{\theta_k}}] - \frac{\partial{(u_k(x;\theta)}}{\partial{\theta_k}})({\bf E_{\theta}}[\frac{\partial{u_l(x;\theta)}}{\partial{\theta_l}}] - \frac{\partial{(u_l(x;\theta)}}{\partial{\theta_l}})].
\end{split}
\end{equation}

With diagonal metric and finite approximation the $k$-th dimension of representation in the linearized tangent space will be 

\begin{equation}
\label{eq:normed_vec}
    \mathcal{V}_k = G_{\theta}^{-\frac 12} V_k \approx G_{kk}^{-\frac 12} V_k =\frac {{\bf E_{\theta}}[\frac{\partial{u_k(x;\theta)}}{\partial{\theta_k}}] - \frac{\partial{(u_k(x;\theta)}}{\partial{\theta_k}}} {{\bf E_{\theta}^{\frac 12}}[({\bf E_{\theta}}[\frac{\partial{u_k(x;\theta)}}{\partial{\theta_k}}] - \frac{\partial{(u_k(x;\theta)}}{\partial{\theta_k}})^2]}.
\end{equation}

Restricted Boltzmann Machines (RBM) \cite{hinton2002training} are special MRFs with maximal cliques of size two and linear clique potentials of the visible and hidden units therefore the Fisher information can be approximated by the covariance of the partial derivatives of the potential functions with expectation taken over a known set of samples. 

\subsection{Outer product metric of discriminative ReLU networks with sparse gradient} 

Since our models are discriminative and not generative, the loss surfaces are not known in absence of the labels.  Hence we define GradNet, a multi-layer network  over the tangent space, as  $h_{\mbox{\scriptsize GradNet}}(x;l(x,\hat{c};\theta))$ (see Fig.~\ref{fig:gradnet})  with the assumption  that the final output of the network after training is $\argmax_{c} \sum_{\hat{c}} h_{\mbox{\scriptsize GradNet}}(x;l(x;\hat{c},\theta)$. 

Results in \cite{denil2013predicting,denton2014exploiting,choromanska2015loss} indicate high over-parametrization and redundancy in the parameter space, especially in deeper feedforward networks, therefore the outer product structure is highly blocked particularly in case of ReLU networks and sparsified gradients. 

\begin{figure}
\caption{Important edges in the gradient graph of the MNIST network.}
\centerline{\includegraphics[scale=.4]{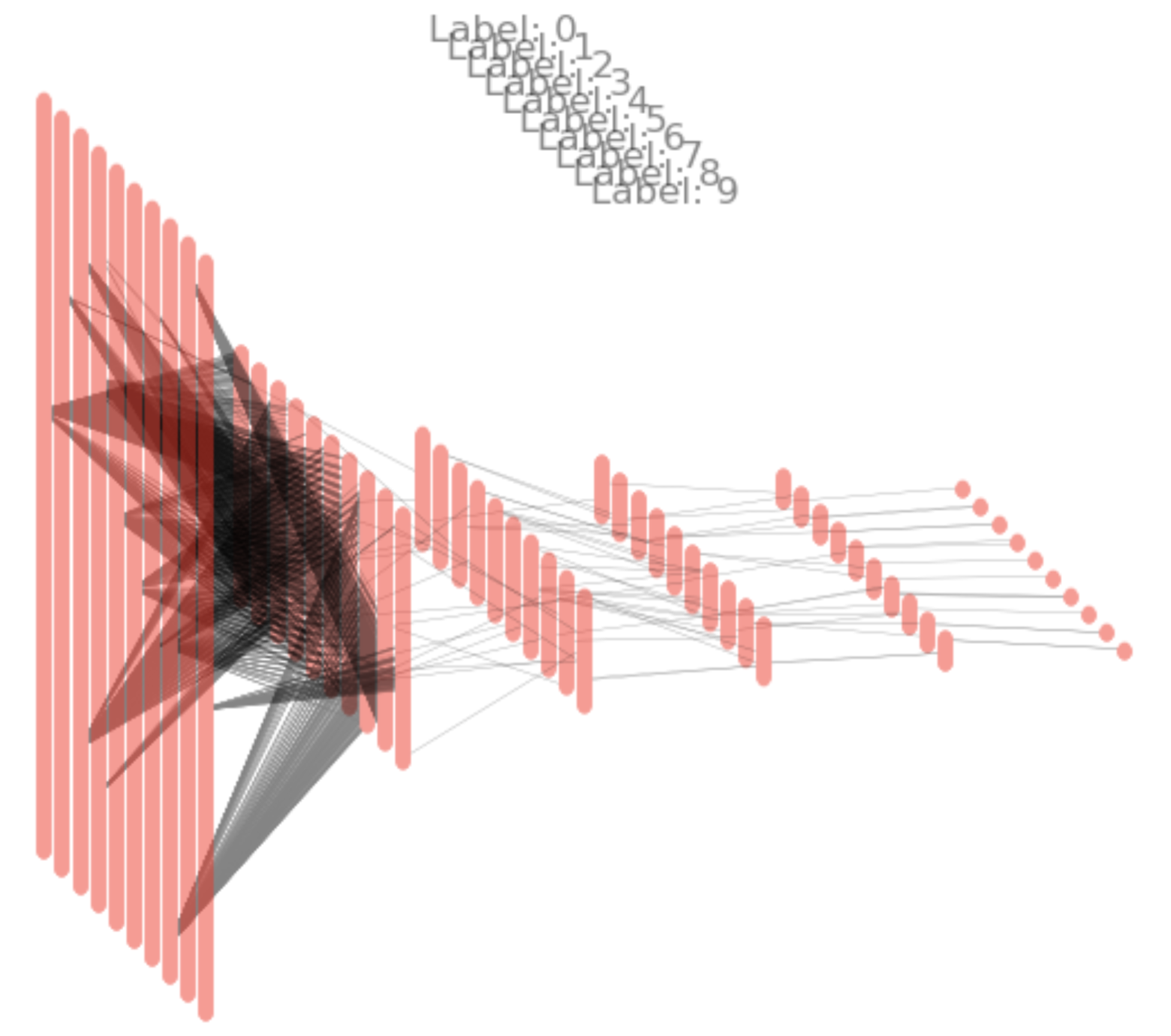}}
\label{fig:grad1}
\end{figure}

Let us consider a multi-layer perceptron with rectified linear units and a gradient graph with sparsity factor $\alpha$ corresponding to the proportion of the most important edges in the gradient graph derived by the Hessian for a particular sample. The nodes are corresponding to the parameters and we connect two nodes with a weighted edge if their value in the Hessian matrix is nonzero and their layers are neighbors. We sparsify the gradient graph layer-by-layer by the absolute weight of the edges. As shown in Fig.~\ref{fig:grad1}, the resulting sparse graph describes the structure of the MINST task well: for each label 0--9, a few nodes are selected in the first layer and only a few paths leave the second layer. 

\section{Experiments}

In our first experiment we trained RBM models with $16$ and $64$ hidden units on the first half of the MNIST \cite{lecun1998gradient} training set and calculated the normalized gradient vectors as in (\eqref{eq:normed_vec}). We used the RBMs output and the normalized gradient vectors within a linear model. The results in Table~\ref{tab:rbm} show that the normalized gradient vector space performed very similar after some initialization with $1k$ sample and after training while the original latent space performed poorly immediately after initialization. 

\begin{figure}
\centering
\caption{GradNet}
\includegraphics[width = .65\textwidth]{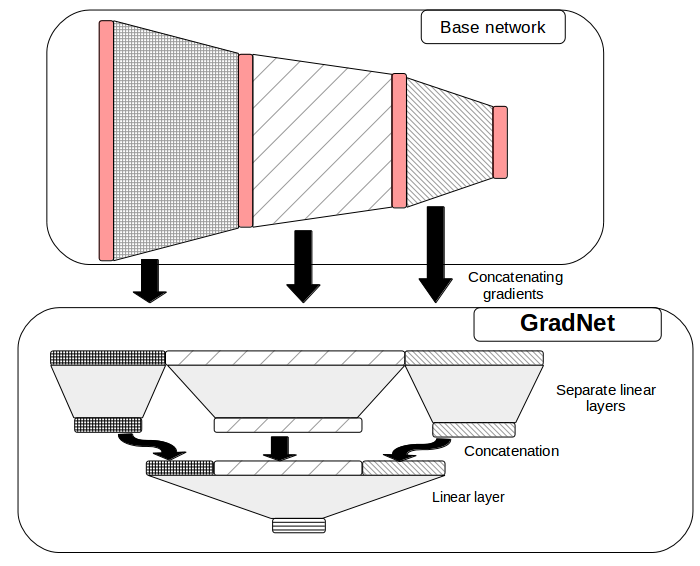}
\label{fig:gradnet}
\end{figure}

We assess the expressive power of the outer product manifold by inspecting the gradient-sets of several CNN's and MLP's. In our experiments we test the hypothesis that for a given CDFNN pre-trained for a classification problem, a quadratic separator on the gradient space can outperform the original ("base") network. We wish to find the proper separator by training a two-layer NN. In order to avoid having too much parameters to train, we chose the hidden layer of our network to have a block-like structure demonstrated in Figure \ref{fig:gradnet}. This model is capable of capturing connections between gradients from adjacent layers of the base network.

In order to find the optimal architecture, the right normalization process and regularization technique, and the best optimization method, we experimented with a large number of setups. We measured the performance of these various kinds of GradNet models on the gradient space of a CNN trained on the first half of the CIFAR-10 \cite{krizhevsky2009learning} training dataset. We used the other half of the dataset and random labels to generate gradient vectors to be used as training input for the GradNet. In the testing phase we use all of the gradient vectors for every data point in the test set, we give them all to the network as inputs, and we define the prediction as the index of the maximal element in the sum of the outputs.

\begin{algorithm}[H]
\caption{Training procedure of GradNet}
\textbf{Input:} Pre-trained model with parameter $\theta$, dataset $D$, GradNet $N$, normalization method $n$, number of epochs $t$

\textbf{Output:} Trained GradNet

\begin{algorithmic}[1]
\Procedure{Train}{$M, D, N, n, t$}

\For{\texttt{epoch from 1 to t}}
\For{\texttt{$batch$ in $D$}}
\State $X \gets augmentation(batch)$
\State $c \gets$ real labels for each data point in $batch$
\State $\hat{c} \gets$ random labels for each data point in $batch$ 
\State $X_g \gets \nabla_{\theta}(l(x;\hat{c},\theta))$ for each data point in the $batch$
\State $\hat{X_g} \gets n(X_g)$  \Comment{normalization}
\State $N \gets update(N, \hat{X_g}, c)$ \Comment{update network with normalized gradients}
\EndFor
\EndFor
\State \textbf{return} $N$\Comment{Return trained $N$}

Prediction for data point $x$: $\argmax_{c} \sum_{\hat{c}} N(n(\nabla_{\theta}l(x;\hat{c},\theta)))$


\EndProcedure
\end{algorithmic}
\end{algorithm}

During our experiments, as a starting point we stopped the underlying original CNN at $0.72$ accuraccy and compared the following settings. 
\begin{itemize}
\item Regarding \underline{\smash{regularization}}, we considered using dropout, batch normalization, both of them together, or none.
\item We experimented with SGD and Adam \underline{\smash{optimization methods}}.
\item Since we suspected that not all coordinates of the gradients are equally important, we only used the elements of large absolute value making the process computationally less expensive. We kept the elements of absolute value greater than the \underline{\smash{$q$-th percentile}} of the absolute value vector, and we tested our model setting this $q$ value for 99, 95, 90, 85, 80 and 70. We also tried a method where we pre-computed the indices of the most important 10$\%$ of the values for each label, and used this together with the above technique.
\item In order to determine the exact \underline{structure} of the GradNet, we tried layers and blocks of different sizes. These models differ only in the size and partition of the middle layer, which were the following in our tests: 5+25+10; 20+100+40; 10+50+20; 5+100+25; 10+200+50; 20+400+100. 
\item In terms of \underline{normalization}, we ran tests using standard norm with and without L2-norm following it; scale norm; power norm with exponents $\frac{1}{8}$, $\frac{1}{4}$, $\frac{1}{2}$, and $2$; and scale norm followed by power norm with exponent $\frac{1}{2}$.
\end{itemize}

\begin{figure}
\begin{minipage}[c]{0.5\linewidth}
\caption{Optimization methods}
\includegraphics[width = \textwidth]{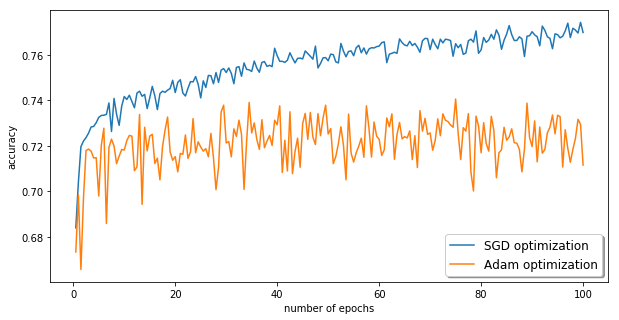}
\label{fig:opt}
\end{minipage}
\hfill
\begin{minipage}[c]{0.5\linewidth}
\caption{Regularization methods}
\includegraphics[width=\textwidth]{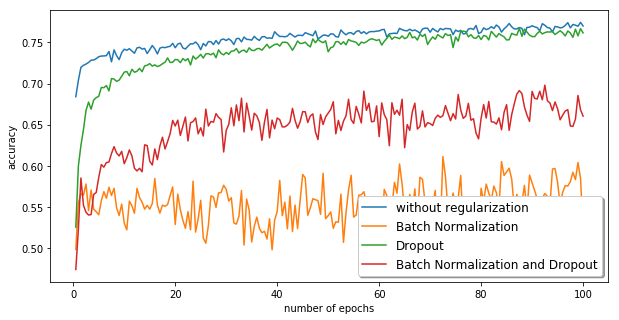}
\label{fig:reg}
\end{minipage}
\end{figure}

\begin{figure}
\begin{minipage}[c]{0.5\linewidth}
\caption{Selection percentile}
\includegraphics[width = \textwidth]{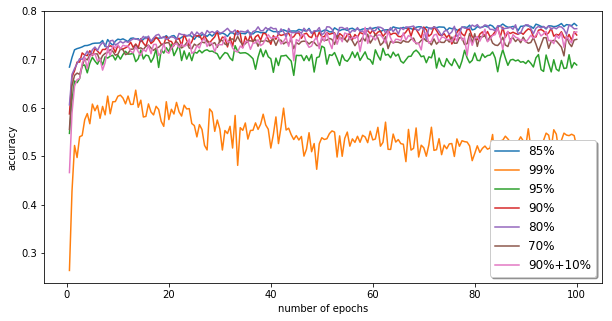}
\label{fig:percent}
\end{minipage}
\hfill
\begin{minipage}[c]{0.5\linewidth}
\caption{Structure}
\includegraphics[width=\textwidth]{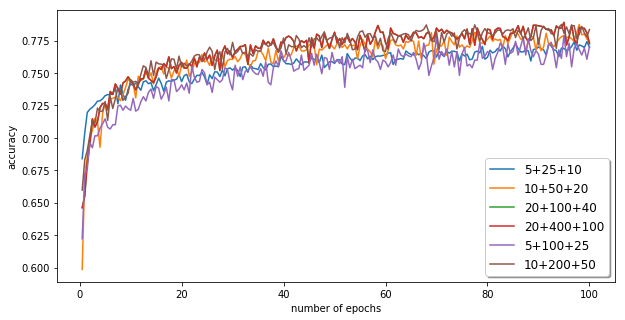}
\label{fig:structure}
\end{minipage}
\end{figure}

\begin{figure}
\begin{minipage}[c]{0.5\linewidth}
\caption{Normalization with structure 5+25+10}
\includegraphics[width = \textwidth]{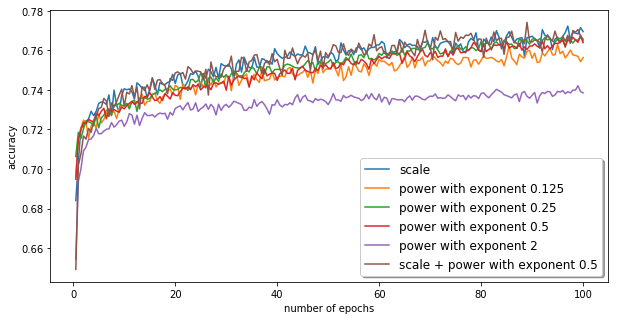}
\label{fig:norm1}
\end{minipage}
\hfill
\begin{minipage}[c]{0.5\linewidth}
\caption{Normalization with structure 5+100+25}
\includegraphics[width=\textwidth]{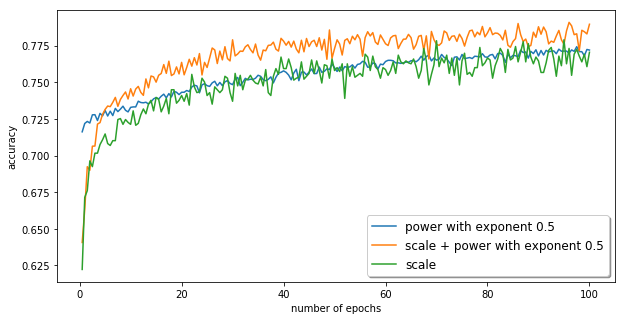}
\label{fig:norm2}
\end{minipage}
\end{figure}

  Learning curves for the different networks are presented in Figures \ref{fig:opt} - \ref{fig:norm2}. We observed that SGD gives a better performance than Adam (Fig.~\ref{fig:opt}), and that regularization is not needed (Fig.~\ref{fig:reg}). We also found that it is sufficient to use the elements of each gradient vector that are greater than the 85-th percentile of all of the absolute values in the vector (Fig.~\ref{fig:percent}). Regarding structure, the best-performing GradNet was the one with hidden layer of size 130 partitioned into sublayers of sizes 5, 100 and 25 (Fig.~\ref{fig:structure}). Interestingly, the GradNet surprassed the performance of the underlying CNN, at some settings even after only one epoch. Out of all the considered normalization methods, the scale norm and the power norm together gave the most satisfactory outcome (Fig.~\ref{fig:norm1},\ref{fig:norm2}). 

\begin{table}
    \centering
    \label{tab:rbm}
    \caption{Performance measure of the normalized gradient based on RBM.}
\begin{tabular}{| c | c | c |}
    \hline
    \multicolumn{3}{|c|}{MNIST}\\
    \hline
    \#hidden & Original & Improved \\ \hline
    16  & 0.6834 & 0.9675 \\ \hline
    16 & 0.8997 & 0.9734 \\ \hline
	64 & 0.872 & 0.9822 \\ \hline
	64 & 0.9134 & 0.9876 \\ \hline
    \end{tabular}
\end{table}

To show the performance of the GradNet with these particular settings, we took snapshots of a CNN at progressively increasing levels of pre-training, and we trained the GradNet on the gradient sets of these networks. We ran these tests using a CNN trained on half of the CIFAR dataset and with one trained on half of MNIST. Table \ref{table:eval} shows the accuracies of all the base networks together with the accuracies of the corresponding GradNets. 

\begin{table}
    \centering
    \caption{Performance measure of the improved networks.}
    \begin{tabular}{| c | c | c |}
    \hline
    \multicolumn{3}{|c|}{CIFAR}\\
    \hline
    Original & Improved & Gain \\ \hline
    0.79  & 0.8289 & +4.9\% \\ \hline
    0.76 & 0.8201 & +7.9\% \\ \hline
	0.74 & 0.8066 & +9\% \\ \hline
	0.72 & 0.7936 & +10.2\% \\ \hline
	0.68 & 0.7649 & +12.5\% \\ \hline
	0.65 & 0.7511 & +15.5\% \\ \hline
	0.62 & 0.7274 & +17.3\% \\ \hline
	0.55 & 0.7016 & +27.5\% \\ \hline
	0.51 & 0.6856 & +34.4\% \\ \hline
	0.49 & 0.678 & +38.3\% \\ \hline
    \end{tabular}
    \begin{tabular}{| c | c | c |}
    \hline
    \multicolumn{3}{|c|}{MNIST}\\
    \hline
    Original & Improved & Gain \\ \hline
    0.92  & 0.98 & +6.5\% \\ \hline
    0.96 & 0.9857 & +2.7\% \\ \hline
	0.9894 & 0.9914 & +0.2\% \\ \hline
    \end{tabular}
    
    \vspace{5 pt}
    \label{table:eval}

\end{table}

\section{Conclusions}

In this paper we proposed a class of reparametrization invariant metrics for discriminative feed-forward networks based on the underlying Riemannian structure. By approximation of the inner product, we showed promising results with our GradNet network in the sparsified gradient space. GradNet outperformed the original network even if built from a few epochs of the original network. We proposed a closed approximation for Restricted Boltzmann Machines and outer product metrics. Our results show high invariance to reparametrization in RBM. In the future, we would like to extend our method to Hessian metrics and further investigate sparsity and possible transitions to less complex manifolds via pushforward and random orthogonal transformations. 

\bibliographystyle{apalike}
\bibliography{hessian}

\end{document}